\documentclass[runningheads]{llncs}
\usepackage{marvosym}
\usepackage[T1]{fontenc}
\usepackage{graphicx}
\usepackage{makecell}
\usepackage{amsmath}
\begin{document}
\title{White Matter Tracts are Point Clouds: Neuropsychological Score Prediction and Critical Region Localization via Geometric Deep Learning\thanks{We acknowledge funding provided by the following National Institutes of Health (NIH) grants: R01MH125860, R01MH119222, R01MH074794, and P41EB015902.}}
\titlerunning{White Matter Tracts are Point Clouds}
%

\author{Yuqian Chen\inst{1,2} \and
Fan Zhang\inst{1} \and
Chaoyi Zhang\inst{2} \and
Tengfei Xue\inst{1,2} \and
Leo R. Zekelman\inst{1} \and
Jianzhong He\inst{4} \and
Yang Song\inst{3} \and
Nikos Makris\inst{1} \and
Yogesh Rathi\inst{1} \and
Alexandra J. Golby\inst{1} \and
Weidong Cai\inst{2} \and
Lauren J. O’Donnell\inst{1}$^($\textsuperscript{\Letter}$^)$
}

\authorrunning{Y. Chen et al.}
\institute{Harvard Medical School, MA, USA \\ \email{odonnell@bwh.harvard.edu} \and The University of Sydney, NSW, Australia  \and The University of New South Wales, NSW, Australia \and  Zhejiang University of Technology, Zhejiang, China}
%
%
%
\maketitle              
\begin{abstract}
White matter tract microstructure has been shown to influence neuropsychological scores of cognitive performance. However, prediction of these scores from white matter tract data has not been attempted. In this paper, we propose a deep-learning-based framework for neuropsychological score prediction using microstructure measurements estimated from diffusion magnetic resonance imaging (dMRI) tractography, focusing on predicting performance on a receptive vocabulary assessment task based on a critical fiber tract for language, the arcuate fasciculus (AF). We directly utilize information from all points in a fiber tract, without the need to average data along the fiber as is traditionally required by diffusion MRI tractometry methods. Specifically, we represent the AF as a point cloud with microstructure measurements at each point, enabling adoption of point-based neural networks. We improve prediction performance with the proposed Paired-Siamese Loss that utilizes information about differences between continuous neuropsychological scores. Finally, we propose a Critical Region Localization (CRL) algorithm to localize informative anatomical regions containing points with strong contributions to the prediction results. Our method is evaluated on data from 806 subjects from the Human Connectome Project dataset. Results demonstrate superior neuropsychological score prediction performance compared to baseline methods. We discover that critical regions in the AF are strikingly consistent across subjects, with the highest number of strongly contributing points located in frontal cortical regions (i.e., the rostral middle frontal, pars opercularis, and pars triangularis), which are strongly implicated as critical areas for language processes.

\keywords{White matter tract  \and Neuropsychological assessment score \and Point cloud \and Deep learning \and Region localization.}
\end{abstract}
\section{Introduction}
The structural network of the brain’s white matter pathways has a strong but incompletely understood relationship to brain function \cite{suarez2020linking,sarwar2021structure}. To better understand how the brain’s structure relates to function, one recent avenue of research investigates how neuropsychological scores of cognitive performance relate to features from diffusion magnetic resonance imaging (dMRI) data. It was recently shown that structural neuroimaging modalities (such as dMRI) strongly contribute to prediction of cognitive performance measures, where measures related to language and reading were among the most successfully predicted \cite{gong2021phenotype}. Other works have shown that measurements of white matter tract microstructure derived from diffusion MRI tractography relate to neuropsychological measures of language performance \cite{zekelman2022white,yeatman2011anatomical}. However, we believe that there are no studies to predict individual cognitive performance based on microstructure measurements of the white matter fiber tracts. Performing accurate predictions of neuropsychological performance could help improve our understanding of the function of the brain’s unique white matter architecture. Furthermore, localizing regions along fiber tracts that are important in the prediction of cognitive performance could be used to understand how specific regions of a white matter tract differentially contribute to various cognitive processes.

\begin{figure}[!t]\centering 
\includegraphics[width=12cm]{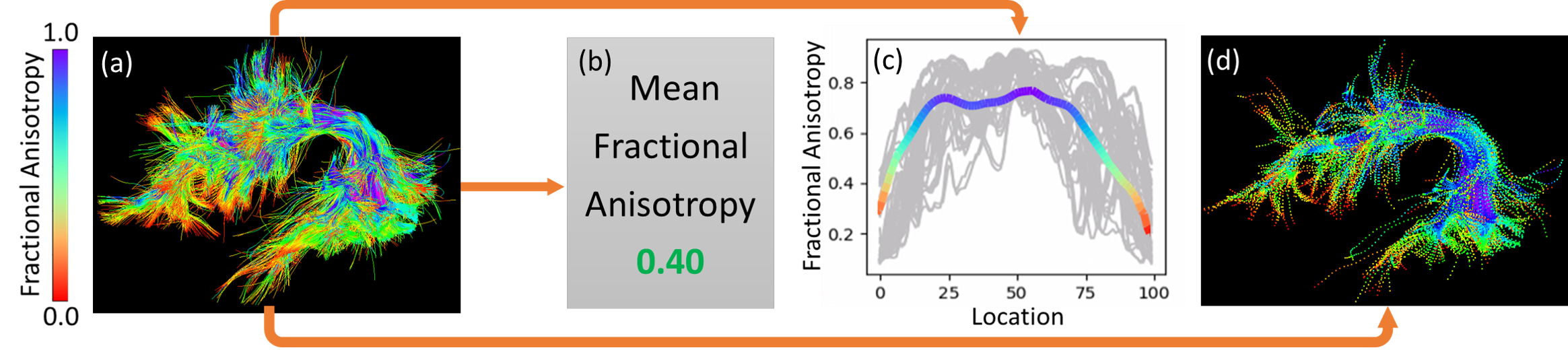}
\caption{(a) Three ways to represent a white matter tract and its microstructure measurements.  (b) Representation with a single mean value; (c) Along-tract representation; (d) Representation with a point cloud.} \label{fig1}
\end{figure}

One important challenge in the computational analysis of white matter fiber tracts is how to represent tracts and their tissue microstructure. Tractography of one anatomical white matter tract, such as the arcuate fasciculus (Fig. \ref{fig1}a), may include thousands of streamlines (or “fibers”), where each streamline is composed of a sequence of points. One tract can therefore contain several hundred thousand points, where each point has one or more corresponding microstructure measurement values. How to effectively represent measurements from points within the tract has been a challenge. One typical approach is to calculate the mean value of measurements from all points within the tract (Fig. \ref{fig1}b). However, this ignores the known spatial variation of measurements along fiber tracts such as the arcuate fasciculus \cite{yeatman2011anatomical,o2009tract}. An alternative is to perform “tractometry” or along-tract analysis (Fig. 1c) to investigate the distribution of the microstructure measures along the fiber pathway \cite{o2009tract,yeatman2012tract}. This method preserves information about the average diffusion measurements at different locations along the tract and thus enables along-tract critical region localization; however, fiber-specific or point-wise information is still obscured.

In contrast to these traditional representations, we propose to represent a complete white matter tract with its set of raw points for microstructure analysis using a point cloud (Fig. \ref{fig1}d). Point cloud is an important type of geometric data structure. In recent years, deep learning has demonstrated superior performance in computer vision tasks \cite{voulodimos2018deep}. Point-based neural networks have demonstrated successful applications on processing geometric data \cite{qi2017pointnet,chen2017multi}. PointNet \cite{qi2017pointnet} is a pioneering effort that enables direct processing of point clouds with deep neural networks. It is a simple, fast and effective framework for point cloud classification and segmentation and has been adapted to various point cloud-related tasks \cite{qi2017pointnet++,zhang2021exploiting,yu20213d}. In previous neuroimaging studies, point cloud representations have been applied to tractography-related learning tasks such as tractography segmentation and tractogram filtering \cite{astolfi2020tractogram,xue2022supwma}. However, to our knowledge, no previous studies have investigated the effectiveness of representing a whole white matter tract and its microstructure measurements using a point cloud for learning tasks. By formulating white matter tract analysis in a point-based fashion using point clouds, we can directly utilize tissue microstructure and positional information from all points within a fiber tract and avoid the need for along-tract feature extraction.

In this study, we propose a framework for neuropsychological assessment score prediction using pointwise diffusion microstructure measurements from white matter tracts. The framework is a supervised deep learning pipeline that performs a regression task. This paper has four contributions. First, we represent white matter tracts as point clouds to preserve diffusion measurement information from all fiber points. To our knowledge, this is the first investigation to study microstructure measurements within white matter tracts using point clouds for a learning task. Second, for regression, we propose the Paired-Siamese Loss to utilize information about the differences between continuous labels, which are neuropsychological scores in our study. Third, our framework is able to perform the task of critical region localization. We propose a Critical Region Localization (CRL) algorithm to recognize critical regions within the tract where points that make important contributions to the prediction task are located. Fourth, using data from 806 subjects from the Human Connectome Project, our approach demonstrates superior performance for neuropsychological score prediction via evaluations on a large-scale dataset and effectiveness in predictively localizing critical language regions along the white matter tract. 

\section{Methods}
\subsection{dMRI Dataset, Neuropsychological Assessment, and Tractography}
To evaluate performance of our proposed method, this study utilizes neuropsychological assessment and dMRI data from the Human Connectome Project (HCP), a large multimodal dataset composed of healthy young adults (downloaded from \textit{http://www.humanconnectomeproject.org/}) \cite{van2013wu}. We study data from 806 unrelated subjects (380 males and 426 females, aged $28.6 \pm 3.7$) with 644 (80\%) for training and 162 (20\%) for testing. Scores from the NIH Toolbox Picture Vocabulary Test (TPVT) are investigated. This computerized neuropsychological assessment measures receptive vocabulary and semantic memory, key components of language functioning \cite{gershon2013iv}. TPVT scores range from 90.69 to 153.09 in this subject cohort. The investigated dMRI data comes from the HCP minimally preprocessed dataset (b = 1000, 2000 and 3000 s/$mm^2$, TE/TR =89/5520 ms, resolution = 1.25 × 1.25 × 1.25 $mm^3$) \cite{van2013wu}. For each subject, the b = 3000 shell of 90 gradient directions and all b = 0 scans are extracted for tractography because this single shell can reduce computation time and memory usage while providing the highest angular resolution for tractography [18,19]. Whole brain tractography is generated from each subject’s diffusion MRI data using a two-tensor unscented Kalman filter (UKF) method \cite{C30,o2012unbiased}. White matter tracts (left AF tract in our study) are identified with a robust machine learning approach that has been shown to consistently identify white matter tracts across the human lifespan and is implemented with the WMA package \cite{zhang2018anatomically}. For each individual’s tract, two measurements are calculated: the tract-specific fractional anisotropy (FA) of the first tensor, which corresponds to the tract being traced, and the number of streamlines (NoS). These measurements of the left AF tract have been found to be significantly related to TPVT scores \cite{zekelman2022white}. Tracts are visualized in 3D Slicer via the SlicerDMRI project \cite{norton2017slicerdmri}.

\subsection{Point Cloud Construction}
The left arcuate fasciculus (AF) tract is selected for the task of predicting TPVT scores because TPVT performance was shown to be significantly associated with left AF microstructure measurements \cite{zekelman2022white}. Each AF tract is represented as a point cloud and N points are randomly sampled from the point cloud as input. Considering the large number of points within the AF tract (on average approximately 300,000 per subject), random sampling serves as a natural but effective data augmentation strategy. Each point has 5 input channels including its spatial coordinates as well as white matter tract measurements (FA and NoS). Therefore, a N×5 point cloud forms the input of the neural network.

\subsection{Network Architecture and Proposed Paired-Siamese Loss}
\begin{figure}[!t]\centering 
\includegraphics[width=12cm]{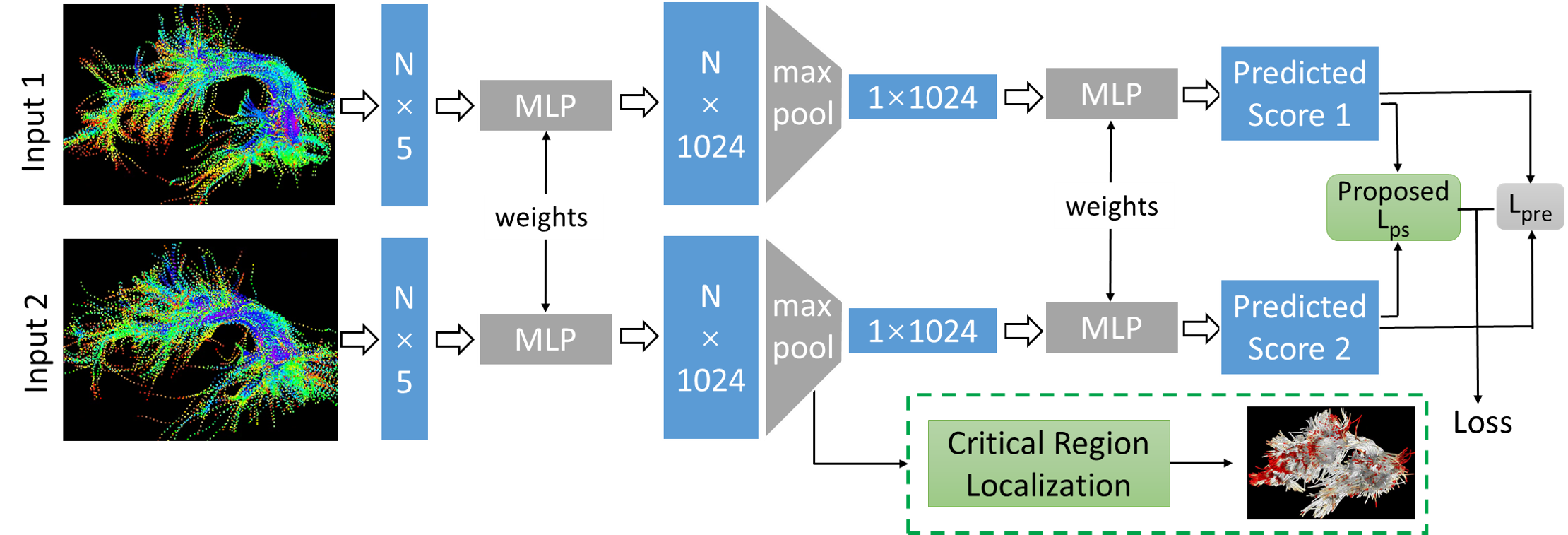}
\caption{Overall pipeline. A pair of AF tracts are represented as a pair of point clouds and fed into a point-based Siamese Network. During training, predicted TPVT scores are obtained from the network and a prediction loss ($L_{pre}$) is calculated as the mean of MSE losses between the predicted scores and ground truth scores. Paired-Siamese loss ($L_{ps}$) is calculated as the MSE between two differences: the difference between predicted scores and the difference between ground truth scores of input pairs. $L_{ps}$ is added to the total loss with weight $w$. During inference, critical regions are recognized with our proposed CRL algorithm, as shown in the green dashed box.} \label{fig2}
\end{figure}

In this work, we propose a novel deep learning framework (Fig. \ref{fig2}) that performs a regression task for TPVT score prediction. The most important aspect of our framework’s design is the supervision of neural network backpropagation using information from the difference of continuous labels (TPVT scores). Unlike a classification task where different categories of labels are independent from each other, a regression task has continuous labels as ground truth. We hypothesize that the relationships between inputs with continuous labels can be used to improve the performance of regression.

For this goal, we propose to adopt a Siamese Network \cite{chopra2005learning} that contains two subnetworks with shared weights. The subnetwork of our Siamese Network is developed from PointNet \cite{qi2017pointnet}. To perform the regression task, the output dimension of the last linear layer is set to 1 to obtain one predicted score. T-Net (the spatial transformation layer) is removed from PointNet to preserve anatomically informative information about the spatial position of tracts \cite{xue2022supwma}. During training, a pair of point cloud sets are used as input to the neural network, and a pair of predicted TPVT scores as well as the difference between them are obtained from the network. For inference, one subnet of the Siamese Network is retained with one point cloud as input.

A new loss function, the Paired-Siamese Loss ($L_{ps}$), is proposed to guide the prediction of TPVT scores. It takes the relationship between the input pair into consideration when training the model. $L_{ps}$ is defined as the mean squared error (MSE) between two differences, where the first is the difference between predicted TPVT scores of an input pair and the second is the difference between ground truth scores of an input pair, as follows:

\begin{equation}
\mathit{L_{ps}}= \frac{1}{N}\sum_{i}((y_{i1}-y_{i2})-(\hat{y_{i1}}-\hat{y_{i2}}))^{2}
\end{equation}

\noindent where $y_{i1}$ and $y_{i2}$ are labels of the input pair and $\hat{y_{i1}}$ and $\hat{y_{i2}}$ are predicted scores of the input pair. In addition to the proposed difference loss, the prediction losses ($L_{pre1}$ and $L_{pre2}$) of each input are calculated as the MSE loss between the predicted score and ground truth score. Then the overall prediction loss of the input pair, $L_{pre}$, is the mean of $L_{pre1}$ and $L_{pre2}$. Therefore, the total loss of our network is $Loss$ =$L_{pre}$+$w$$L_{ps}$, where $w$ is the weight for $L_{ps}$.

\subsection{Critical Region Localization} 
We propose a Critical Region Localization (CRL) algorithm to identify critical regions along fiber tracts that are important in the prediction of cognitive performance. The algorithm steps are explained in detail as follows. First, to ensure that critical points are identified from the whole point set of the AF tract, we divide the tract into multiple point sets (N points per set) without replacement and sequentially feed these point sets to the network. Second, we propose a Contributing Point Selection (CPS) module to obtain critical points and their weights for each input point set. In CPS, we intuitively take advantage of the max-pooling operation in the network as in \cite{qi2017pointnet} to obtain point sets that contribute to the max-pooled features (referred to as contributing points in our algorithm). However, instead of taking contributing points directly as critical points as in \cite{qi2017pointnet}, our method considers the number of max-pooled features of each contributing point and assigns this number to the point as its weight. The lists of contributing points and their weights are then concatenated across all point sets. Third, the previous two steps are repeated M times (so that every point in the AF tract is fed into the network M times for CPS). The weights of each contributing point are summed across all repetitions. Finally, we define critical regions along the AF tract as the points that have top-ranking weights for each subject (the top 5\% of points are chosen in this paper to identify highly important regions). These regions are then interpreted in terms of their anatomical localization by identifying the Freesurfer \cite{fischl2012freesurfer} regions in which they are located.

\subsection{Implementation Details} 
In the training stage, our model is trained for 500 epochs with a learning rate of 0.001. The batchsize of training is 32 and Admax \cite{kingma2014adam} is used for optimization. We use a weight decay with the corresponding coefficient 0.005 to avoid overfitting. All experiments are performed on a NVIDIA RTX 2080 Ti GPU using Pytorch (v1.7.1) \cite{paszke2019pytorch}. The weight of difference loss $w$ is empirically set to be 0.1 \cite{wang2020anterior,fu2020domain}. The number of input points N is set to be 2048 considering memory limitation. On average, each epoch (training and validation) takes ~4 seconds with 2GB GPU memory usage. Number of iteration M in the CRL algorithm is set to 10.

\section{Experiments and Results}
\subsection{Evaluation Metrics} 
Two evaluation metrics were adopted to quantify performance of our method and enable comparisons among approaches. The first metric is the mean absolute error (MAE), which is calculated as the averaged absolute value of the difference between the predicted TPVT score and the ground truth TPVT score across testing subjects. The second metric is the Pearson correlation coefficient ($r$), which measures the linear correlation between the predicted and ground truth scores and has been widely applied to evaluating performance of neurocognitive score prediction \cite{gong2021phenotype,kim2021structural,tian2021machine}.

\begin{table}\centering
\caption{Quantitative  comparison results.}\label{tab1}
\begin{tabular}{c|c|c|c|c|c|c|c|c|c}
\hline
Methods & \makecell[c]{Mean+\\LR} & \makecell[c]{Mean+\\ENR} & \makecell[c]{Mean+\\RF} & \makecell[c]{AFQ+\\LR} & \makecell[c]{AFQ+\\ENR} & \makecell[c]{AFQ+\\RF} & \makecell[c]{AFQ+\\1D-CNN} & \makecell[c]{Ours \\ w/o $L_{ps}$ }& Ours\\ 
\hline
\makecell[c]{Input\\ Features} &  \makecell[c]{mean \\FA,\\NoS} & \makecell[c]{mean \\FA,\\NoS} & \makecell[c]{mean \\FA,\\NoS} & \makecell[c]{along-\\tract\\FA,\\NoS} & \makecell[c]{along-\\tract\\FA,\\NoS} & \makecell[c]{along-\\tract\\FA,\\NoS} & \makecell[c]{along-\\tract\\FA,\\NoS} & \makecell[c]{point-\\wise\\FA,\\NoS} & \makecell[c]{point-\\wise\\FA,\\NoS}\\
\hline
MAE & \makecell[c]{6.546\\(4.937)} & \makecell[c]{6.548\\(4.955)} & \makecell[c]{6.58\\(4.812)} & \makecell[c]{7.364\\(5.744)} & \makecell[c]{6.928\\(5.233)} & \makecell[c]{7.034\\(5.447)} & \makecell[c]{ 6.537\\(4.975)} & \makecell[c]{ 6.639\\(5.174)} & \makecell[c]{\textbf{6.512}\\\textbf{(5.081)}}\\ 
\hline
$r$ & 0.062 & 0.089 &0.134 &0.135 &0.225 &0.138 &0.114 &0.316 &\textbf{0.361}\\
\hline
\end{tabular}
\end{table}

\subsection{Evaluation Results}
\subsubsection{Comparison with baseline method}
We compared our proposed approach with several baseline methods. For all methods, features of FA and NoS were used for prediction. First, the three representations of FA measurements of the AF tract, namely mean FA, along-tract FA and pointwise FA (Fig. \ref{fig1}) were compared. To analyze the mean FA representation, Linear Regression (LR), Elastic-Net Regression (ENR) and Random Forest (RF) were performed to predict TPVT scores. Next, along-tract FA was obtained with the Automated Fiber Quantification (AFQ) algorithm \cite{yeatman2012tract} implemented in Dipy v1.3.0 \cite{garyfallidis2014dipy}. FA features from 100 locations along the tract were generated from AFQ. FA features and NoS were concatenated into the input feature vector. For along-tract FA, performance of LR, ENR, RF and 1D-CNN models were investigated. We adopted the 1D-CNN model proposed in a recent study that performs the age prediction task with microstructure measurements \cite{he2022model}. The parameters of all methods were fine-turned to obtain the best performance. All results are reported using data from all testing subjects. As shown in Table 1, our proposed method outperforms all baseline methods in terms of both MAE and $r$, where the $r$ value of our method shows an obvious advantage over the other methods. The reason for the largely improved $r$ but slightly improved MAE is that several baseline methods give predictions narrowly distributed around the mean, producing a reasonable MAE but a low $r$. The results suggest that point clouds can better represent microstructure measurements of white matter fiber tracts for learning tasks. 

\begin{figure}[!t]\centering 
\includegraphics[width=11cm]{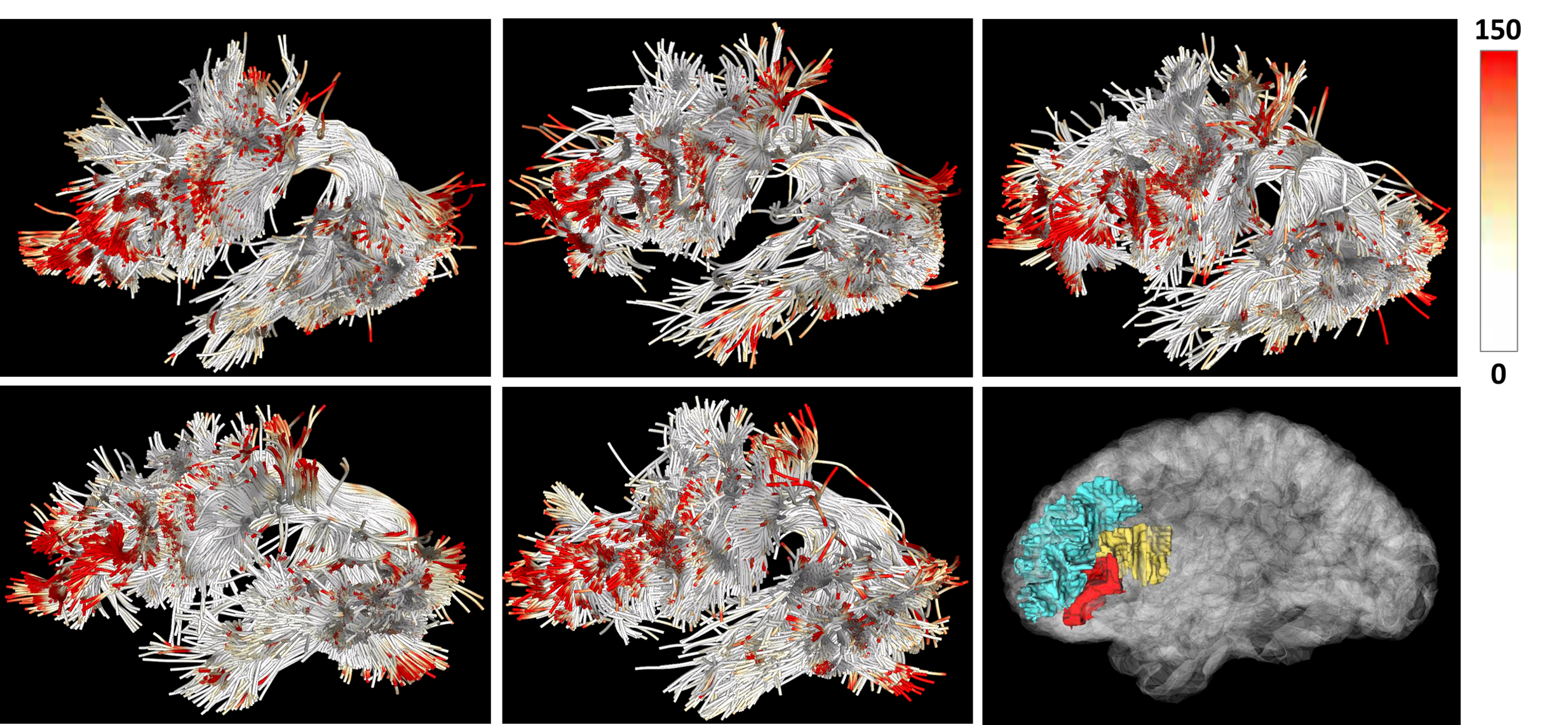}
\caption{The weights for contributing points within the AF tract show strong consistency of critical predictive regions across five randomly selected subjects. Lower right image shows the top three cortical areas intersected by critical AF regions (blue: rostralmiddlefrontal; yellow: parsopercularis; red: parstriangularis).} \label{fig4}
\end{figure}

\subsubsection{Ablation study}
We performed an ablation study to evaluate the effectiveness of our proposed Paired-Siamese Loss. Model performances with (Ours) and without the $L_{ps}$ (Ours w/o $L_{ps}$) were compared. As shown in Table \ref{tab1}, adding the proposed loss obviously improved performance, indicated by a larger $r$ value and a smaller MAE. The large improvement of $r$ is likely because the proposed loss takes the relationship between the input pair into consideration during training.

\subsubsection{Interpretation of critical regions}
Critical points and their corresponding importance values (weights) were obtained for all 162 testing subjects during inference by applying our proposed CRL algorithm. The obtained critical regions within the tract are visualized in Fig. \ref{fig4}. It is apparent that the anatomical locations of the critical regions are strikingly consistent across subjects, demonstrating the potential utility of our proposed method for investigating the relationship between local white matter microstructure and neurocognitive assessment performances. For prediction of TPVT, we find that most critical region points are located near the interface of the white matter and the cortex, where 34.6\% are in the white matter and 64.2\% in the cortex, on average across all testing subjects. Investigating the critical regions of the left AF that are important for language performance as measured by TPVT, we find that on average 16.5\% of the critical region points are located in the rostral middle frontal cortex, 9.1\% in the cortex of the pars opercularis, and 7.6\% in the cortex of the pars triangularis. This finding is in line with the understanding that frontal cortical regions are critical for semantic language comprehension and representational memory \cite{dronkers2011neural,goldman2011circuitry}.

\section{Conclusion}
To conclude, we propose a novel deep learning based framework for predicting neuropsychological scores. In our framework, we represent white matter tracts as point clouds, to preserve point information about diffusion measurements and enable efficient processing using point-based deep neural networks. We propose to utilize the relationship between paired inputs with continuous labels during training by adding a Paired-Siamese Loss. In addition, we propose a critical region localization (CRL) algorithm to obtain the importance distribution of tract points and identify critical regions within the tract. Our method outperforms several compared algorithms on a large-scale HCP dataset. We find that points located near the interface between white matter and cortex, and especially in frontal cortical regions, are important for prediction of language performance. Potential future work includes improving our framework by addressing some existing limitations such as ignorance of continuous streamline information and lack of density correction of tractograms.





%
%
%
\bibliographystyle{splncs04}
\bibliography{mybibliography}
%

\end{document}